\documentclass[letterpaper, 10 pt, conference, onecolumn]
{ieeeconf}  

\IEEEoverridecommandlockouts                              

\usepackage[ruled,linesnumbered]{algorithm2e}
\usepackage{balance}
\usepackage{graphicx}
\usepackage[cmex10]{amsmath}
\usepackage{multirow}
\usepackage{booktabs}
\usepackage{diagbox}
\usepackage{amssymb}
\usepackage{helvet}
\usepackage{courier}
\usepackage{slashbox}
\usepackage{latexsym}

\title{\LARGE \bf
Graph Clustering with Cross-View Feature Propagation
}

\author{Zhixuan Duan$^{1}$, Zuo Wang$^{1}$ and Fanghui Bi$^{1}$
\thanks{$^{1}$Zhixuan Duan, Zuo Wang, and Fanghui Bi are with the College of Computer and Information Science, Southwest University, Chongqing 400715, China
        {(Email: \{qqww123456, wz20011013\}@email.swu.edu.cn, bifanghui@cigit.ac.cn)}.}%
}

\begin{document}

\maketitle
\thispagestyle{empty}
\pagestyle{empty}

\begin{abstract}

Graph clustering is a fundamental and challenging learning task, which is conventionally approached by grouping similar vertices based on edge structure and feature similarity.
In contrast to previous methods, in this paper, we investigate how multi-view feature propagation can influence cluster discovery in graph data.
To this end, we present Graph Clustering With Cross-View Feature Propagation (GCCFP), a novel method that leverages multi-view feature propagation to enhance cluster identification in graph data.
GCCFP employs a unified objective function that utilizes graph topology and multi-view vertex features to determine vertex cluster membership, regularized by a module that supports key latent feature propagation. 
We derive an iterative algorithm to optimize this function, prove model convergence within a finite number of iterations, and analyze its computational complexity. 
Our experiments on various real-world graphs demonstrate the superior clustering performance of GCCFP compared to well-established methods, manifesting its effectiveness across different scenarios.

\end{abstract}

\section{Introduction}\label{sec:introduction}

The complex system containing interactive components can generally be modeled as a graph, wherein vertices and edges represent components (data samples) and component-component relationships.
Real-world graph data representing complex systems are ubiquitous. 
Cited documents, such as online documents and scientific articles, can be represented as a document graph describing the intricate citations between documents.
In contrast to factitiously generated graphs, real-world ones always carry latent structures indicating the cohesiveness of vertices, which may provide convenience to further analytical tasks.
Aiming to uncover the latent groups of cohesive vertices, cluster analysis in graph-structured data has been a fundamental and challenging task of data mining and machine learning\cite{hu2023fcan}.

Despite a challenging problem, graph clustering urgently needs solving as it is directly related to several real-world analytical tasks of great importance, such as document classification \cite{kipf2016variational}, biological module identification \cite{he2018measuring}, social community detection \cite{liu2022symmetry}, and link analysis\cite{li2022second,luo2021alternating,li2022momentum}.
Several methods have been proposed to tackle the graph clustering problem effectively.
These approaches can be categorized based on the data type for cluster analysis.

Some approaches can uncover graph clusters by leveraging topological information.
Representatives such as Fast unfolding \cite{blondel2008} and the Clauset-Newman-Moore algorithm \cite{clauset2004} use edge structure to maximize modularity, which quantifies the density difference between purposeful and random vertex groups.
Model-based approaches, such as Community through directed affiliations (CoDA) \cite{yang2014detecting,zeng2024novel}, and Probabilistic stochastic block models (SBMs) \cite{peng2015scalable,wei2022robust,qin2022low}, also can reveal graph clusters from edge structure.
Other typical approaches, including Normalized cut (NCut) \cite{shi2000} and Homophily preserving community detection (HPNMF) \cite{ye2019homophily,bedi2016community}, and matrix factorization \cite{liu2023high,luo2021symmetric,yuan2024fuzzy,wang2024distributed,li2023saliency,zhu2021community,qin2023asynchronous,liu2021edmf} can uncover graph clusters using edge weights\cite{chen2022mnl}, which indicate the local topological similarity between connected vertices.

In contrast to uncovering graph clusters solely utilizing topological information \cite{luo2021adjusting}, many recent approaches to graph clustering can perform the task by considering graph topology and some supplementary information\cite{luo2021fast,luo2022neulft,bi2023proximal,chen2021hierarchical}, e.g., vertex features.
To differentiate from those graphs solely constituting topological information, attributed graph (AG) \cite{he2017misaga} is proposed to make use of graph topology and vertex features to characterize vertices.
In an AG, those groups containing vertices sharing high edge density and similar features are engrossing.

Cluster analysis in attributed graphs (AGs) can generally be achieved using two methodologies.
The first involves in constructing a weighted graph, where edge weights quantify similarity or correlation regarding graph topology and features between vertices.
Clusters in the weighted graph can then be discovered by commonly used clustering approaches, such as spectral \cite{guo2017combining,cheng2021novel,tsitsulin2023graph,bianchi2020spectral,li2021adaptive} and fuzzy clustering \cite{cheng2021novel,he2018discovering,edwards1965method,xing2021learning,hu2021fast}.

The second and more prevalent methodology attempts to build a learning model to learn the latent space shared by vertex features and graph topology, representing the cluster assignment of each vertex.
A number of sophisticated machine learning methods have been utilized for this, such as Relational topic models  \cite{lin2021dynamical,luo2020position} and matrix factorizations \cite{zhong2020momentum,yuan2022kalman,yuan2024fuzzy,yuan2023adaptive,yuan2024adaptive,li2023generalized,baltrunas2011matrix,wang2021large,he2021vivinal}.
Additionally, graph multi-view clustering \cite{wang2019gmc,he2019contextual,he2021multi}, and Co-regularized multi-view spectral clustering \cite{kumar2011co} are effective spectral-based approaches to discovering clusters in AGs\cite{hu2020algorithm,he2024polarized,he2021learning,luo2021novel,zhou2024differentiable}.
More recently, various graph neural networks \cite{wang2022multi,chen2024sdgnn,bi2023two,zhong2021alternating,yang2023highly,chen2023tensor,chen2020efficient,zhou2022auto}, especially graph auto-encoders  \cite{luo2023predicting,bi2023fast}, have been proposed to perform graph clustering tasks by projecting node features into the low-dimensional latent space\cite{luo2021generalized,shi2020large,shi2020large,wu2020advancing,wu2021proportional,bojanowski2017optimizing}.

\begin{figure}
	\centering
	\includegraphics[width=\columnwidth]{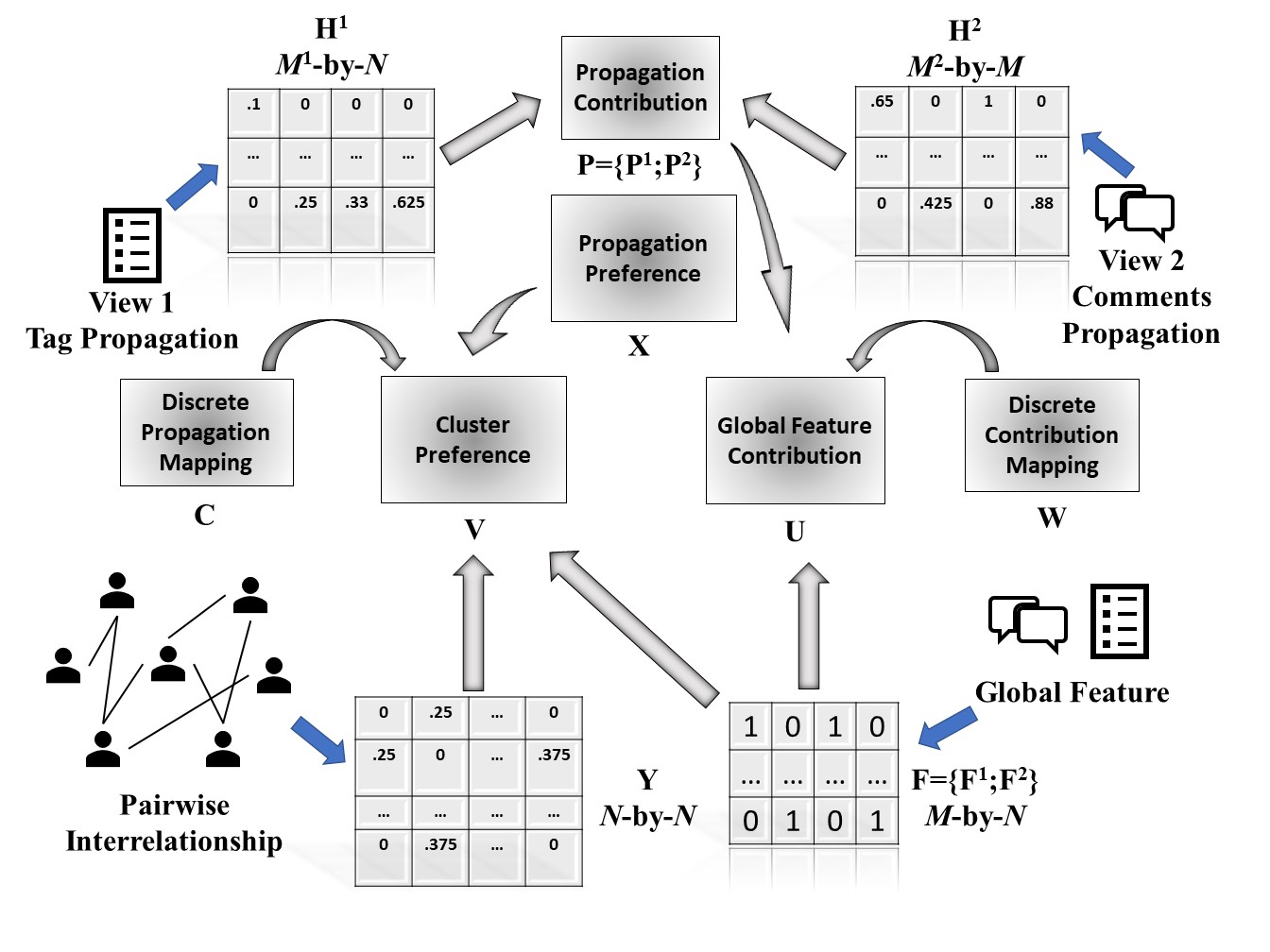}
	\caption{Overall idea of the proposed Graph Clustering With Cross-View Feature Propagation (GCCFP).}
	\label{frame}
\end{figure}

In the Internet era, relational data collected from multiple sources are readily available  \cite{wu2022prediction,wu2020data,li2022diversified,wu2019posterior}.
For example, in a social graph (Fig. \ref{frame}) containing social players (vertices) connected by social ties (edges), additional data such as tags and comments can provide multiple views on vertex features.
These features often propagate among similar social players within social clusters, potentially influencing online behaviors like group formation. 
Given this, it is beneficial to extend traditional attribute graphs (AGs) to Cross-view feature propagation graphs (CVFPGs), incorporating both edge structures and global vertex features concatenated from multi-view features. 
Performing clustering tasks in CVFPG can influence the cluster membership by multi-view feature propagation and infer latent cluster descriptions based on these integrated features\cite{wu2022double,wu2021pid,chen2022differential}.

Although the data are widely available, there is a deficiency in effective approaches that consider edge structure, vertex features, and multi-view feature propagation to jointly impact the cluster membership of each vertex\cite{chen2022growing,wu2023mmlf,li2023nonlinear}.
Moreover, what role multi-view vertex features may play in graph clustering is yet under-explored.
Most existing approaches to graph clustering perform the task by grouping vertices sharing the high similarity of local edge structure and global features, but overlook the potential impact of multi-view feature propagation on the learning of cluster membership for each vertex.
Therefore, it would be greatly beneficial if AG could be appropriately extended to CVFPG, facilitating the investigation of multi-view feature propagation in graph clustering.

To this end, we propose Graph Clustering With Cross-View Feature Propagation (GCCFP), a novel model to solve the mentioned clustering problem in the multi-view feature propagation graph.
GCCFP leverages graph topology and multi-view vertex features to determine vertex cluster membership, which is simultaneously regularized by a module that supports key latent feature propagation.
The proposed GCCFP have been tested on various real-world graphs and compared with well established methods.
The results obtained can demonstrate its superior clustering performance, manifesting its effectiveness across various real-world scenarios.

\section{Pivotal graph clustering with
cross-view feature propagation}\label{method}
In this section, we introduce the details of the proposed GCCFP, starting with the mathematical notations and preliminaries.
We then elaborate on the essential modules constituting GCCFP, including interrelationship modeling and multi-view propagation preserving feature clustering.
With these building blocks, we establish the unified objective function for GCCFP and develop an iterative algorithm for optimizing the unified objective function.

\subsection{Notations}
In this paper, we assume that a graph comprises $N$ vertices and $\mid$E$\mid$ edges.
Additionally, there are $D$ views of vertex features, each view contains $\mathit {M}^i$ features and $\sum_{i=1}^{D} {M}^i = M$. 
To describe such a graph, we use $\mathbf Y \in \{0,1\}^{N \times N}$ and $\mathbf {F} = [\mathbf {F}^1;...;\mathbf {F}^D] \in \{0,1\}^{{M} \times N}$ to represent the vertex adjacency and multi-view vertex features.
Additionally, the non-negative matrix $\mathbf {H}^i \in \mathbb{R}^{+{M}^i \times N}$ represents the feature propagation in view $i$.
GCCFP uses the latent spaces $\mathbf V \in N \times K$ and $\mathbf U \in M \times K$ to represent the cluster membership of each vertex and the global cluster contribution of each feature, respectively.
In addition, GCCFP uses the latent space $\mathbf X \in N \times S$ and the latent space $\mathbf P^i \in M^i \times S$ to respectively represent the shared vertex-propagation preference and view-wise feature-propagation contribution.
The $S \times K$ latent spaces $\mathbf C$ and $\mathbf W$ are utilized by GCCFP to regularize the learning of $\mathbf V$ and $\mathbf U$ to capture the pivotal information on the latent feature propagation.
$\mathbf Y_{ij}$ denotes the $(i,j)$-th element in matrix $\mathbf Y$.
$tr$($\cdot$) and $\Vert$$\cdot$$ \Vert_{F}$ respectively denote the trace operation and the Frobenius norm.

\subsection{Structural propagation modeling}
Modeling graph structure is crucial for cluster analysis in graph data, as it is always seen as the cornerstone of the graph.
The conventional modeling of graph structure can be achieved by assuming the edges in the graph are generated by the cluster membership represented as the latent space.
However, such a modeling method is vulnerable to noisy edges, i.e., ones connecting vertices possessing diverse local structures.
To improve the robustness of the original graph data, we propose the following re-weighting method that is inspired by diffusion theory \cite{coussi1995relation}:
\begin{equation}
	\label{sdiffusion}
	\begin{aligned}
		\mathbf{D}_{ij}\!=\!\begin{cases}
			\frac{(d_i + d_j)[\mathbf {Y}^{T}\mathbf {Y}+\mathbf {Y}]_{ij}}{2d_id_j}&\text{if}\;\mathbf{Y}_{ij}=1,\\
			0&\text{otherwise},
		\end{cases}
		d_i = \sum_{j} \mathbf {Y}_{ij}.
	\end{aligned}
\end{equation}
In cases where two vertices establish a connection, $\mathbf{D}_{ij}$ reveals the probabilities of the bi-directional transfer of the shared structure from one vertex to the other.
We can then use $\mathbf{D}$ obtained by Eq.~(\ref{sdiffusion}) to re-weight the edges in the graph, that is $\mathbf{Y}_{ij}=\mathbf{D}_{ij}\cdot \mathbf{Y}_{ij}$.
The structural propagation of the edges in the graph is therefore integrated into the newly constructed $\mathbf{Y}$.

In this paper, we assume that the structural propagation between pairwise vertices is reconstructed by the product of $\mathbf V$ and $\mathbf V^{T}$. In other words, $\mathbf Y_{ij} = [\mathbf V\mathbf V^{T}]_{ij}$+$\gamma_{ij}$, and $\gamma_{ij}$ represents error.
Thus, the modeling of structural propagation adopted by DMVFPPGC is formulated to minimize the following cost function:
\begin{equation}\label{inter-modeling}
	O_1=\left\Vert\mathbf{Y}-\mathbf V\mathbf V^{T}\right\Vert^{2}_F.
\end{equation}
By minimizing $O_1$ in Eq.~(\ref{inter-modeling}), learning the cluster membership of each vertex in the graph is influenced by the structural propagation carried by $\mathbf Y$.

\subsection{Clustering with cross-view feature propagation}
The propagation of corss-view vertex features may have a view-wise effect on the modeling of graph structure and global features.
To effectively capture such latent effect, for each view of vertex features, GCCFP firstly constructs a matrix representing the propagation of each feature associated with each vertex.
Then, GCCFP tries to learn $\mathbf X$ and $\mathbf P^i$, which represent vertex-propagation preference and feature-propagation contribution, respectively.
Utilizing $\mathbf X$ and $\mathbf P^i$, GCCFP may embed the learning of cluster membership and feature-cluster contribution ($\mathbf V$ and $\mathbf U$) with the latent information on view-wise feature propagation.

\subsubsection{Identifying view-wise feature propagation}
As mentioned above, $\mathbf D$ indicates the structural propagation between connecting vertices.
Given that, we may obtain the view-wise feature propagation as follows:
\begin{equation}\label{feature-propagation}
	\mathbf H^i= \mathbf F^i \mathbf D.
\end{equation}
Given Eq. (\ref{feature-propagation}), we know that $\mathbf {H}^{i}_{jk} = \sum_l \mathbf F^i_{jl}\mathbf D_{lk}$.
In other words, $\mathbf {H}^{i}_{jk}$ equals the frequency of feature $j$ in view $i$ weighted by the propagation of vertex $k$. 
Thus, it quantifies the propagation of feature $j$ caused by vertex $k$.
\subsubsection{Propagation cross-view  feature modeling}
Given $\mathbf H^i$, we assume that it is reconstructed by the product of $\mathbf X^T$ and $\mathbf P^i$, i.e., $\mathbf {H}^{i}_{jk} =[\mathbf P^i\mathbf X^T]_{jk}+\sigma^i_{jk}$, where $\sigma^i_{jk}$ represents the error term.
And we assume that the global feature $\mathbf F$ is reconstructed by the product of $\mathbf V^T$ and $\mathbf U$, i.e., $\mathbf {F}_{jk} =[\mathbf U\mathbf V^T]_{jk}+\phi_{jk}$, where $\phi_{jk}$ represents the error term.
Introducing the discrete matrices $\mathbf C$ and $\mathbf W$, we may formulate the problem of corss-view propagation preserving feature clustering as the minimization of the following cost function:
\begin{equation}\label{feature-clus}
	\begin{aligned}
		&O_2\!=\!\sum_{i}\!\left\Vert \mathbf H^{i}\!-\!\mathbf P^{i}\mathbf X^{T}\right\Vert ^{2}_F\!+\!\left\Vert \mathbf X\mathbf C\!-\!\mathbf V\right\Vert ^{2}_F\!\\
		&+\!\left\Vert \mathbf P\mathbf W\!-\!\mathbf U\right\Vert ^{2}_F+\left\Vert\!\mathbf F- \mathbf U\mathbf V^T\!\right\Vert ^{2}_F,\\
		& \text{subject to }\mathbf P = \lbrack\mathbf P^{1};...;\mathbf P^{D}\rbrack,\mathbf W\mathbf W^T = \mathbf I, \mathbf C\mathbf C^T = \mathbf I.
	\end{aligned}
\end{equation}
Through minimizing the above cost function, GCCFP can obtain the optimal cluster membership $\mathbf V$ and global feature-cluster contributions $\mathbf U$.
Different from previous approaches, $\mathbf V$ and $\mathbf U$ are influenced by the pivotal latent information inferred from corss-view feature propagation.

\subsection{Objective function}
Integrating $O_{1}$ and $O_{2}$ with hyper-parameters and constraints, GCCFP formulates the graph clustering problem as the following optimization problem with unified objective function $O$:
\begin{equation}\label{obj}
	\begin{aligned}
		&\text{minimize}\\
		&O\!=\!\left\Vert\!\mathbf F- \mathbf U\mathbf V^T\!\right\Vert ^{2}_F\!+\!\sum_{i}\!\left\Vert \mathbf H^{i}\!-\!\mathbf P^{i}\mathbf X^{T}\right\Vert ^{2}_F\\
		&+\!\left\Vert \mathbf P\mathbf W\!-\!\mathbf U\right\Vert ^{2}_F+\alpha\left\Vert\mathbf Y-\mathbf V\mathbf V^{T}\right\Vert^{2}_F+\lambda\left\Vert \mathbf X\mathbf C\!-\!\mathbf V\right\Vert ^{2}_F,\\
		&\text{subject to } \mathbf V\!\ge \!0, \mathbf U\!\ge \!0, \mathbf P^{i}\!\ge \!0, \mathbf X\!\ge \!0, \mathbf C\!\ge \!0, \mathbf W\!\ge \!0,\\
		&\mathbf W\mathbf W^T = \mathbf I, \mathbf C\mathbf C^T = \mathbf I,
	\end{aligned}
\end{equation}
where $\alpha$ and $\lambda$ are hyper-parameters for balancing the relative significance of global feature and structural propagation modeling and regularizing multi-view latent propagation in the learning process.
By minimizing $O$, GCCFP can infer the optimal cluster membership ($\mathbf V$) from graph topology and global vertex features. 
Simultaneously, the learning of $\mathbf V$ is influenced by the pivotal latent propagation of multi-view vertex features.

\subsection{Model fitting}\label{opt}
The unified objective function (Eq. (\ref{obj})) proposed in this paper is generally non-convex.
However, it is convex regarding $\mathbf V$, $\mathbf U$,  $\mathbf X$, $\mathbf P^i$, $\mathbf W$, or $\mathbf C$ when other ones are fixed.
Therefore, the model variables in each latent space can be optimized when ones in other latent spaces are fixed. 
We derive the following iterative strategy that updates the latent variables to guide $O$ to converge.

\subsubsection{Optimizing $\mathbf V$}
Denoting $\mathbf{\eta}_{jk}$ as the Lagrange multiplier for $\mathbf{V}_{jk}\ge$ 0, we may construct the Lagrange function of $\mathbf{V}$:
\begin{equation}\label{langran-v}
	L(\mathbf{V},\mathbf{\eta})=O-\mathit{tr}(\mathbf{\eta}^T\mathbf{V}).
\end{equation}
According to the KKT conditions, we can derive the following equation system for obtaining the updating rule of $\mathbf V$:
\begin{equation}\label{kkt-v1}
	\begin{aligned}
		&\frac{\partial L}{\partial \mathbf V_{jk}}=4\alpha [\mathbf{VV}^{T}\mathbf{V}]_{jk} \! - \! 4\alpha[\mathbf{YV}]_{jk}+2\lambda  \mathbf V_{jk}\\
		&+ \! 2[\mathbf V\mathbf U^{T}\mathbf U]_{jk}-2[\mathbf F^{T}\mathbf U]_{jk}-2\lambda [\mathbf{XC}]_{jk}-\eta_{jk}=0,\\
		&\mathbf{\eta}_{jk}\cdot\mathbf{V}_{jk}=0, \mathbf{\eta}_{jk}\ge0, \mathbf{V}_{jk}\ge0.
	\end{aligned}
\end{equation}
The updating rule for $\mathbf V$ can be derived by solving Eq.~(\ref{kkt-v1}):
\begin{equation}\label{rule-v}
	\begin{aligned}
		&\mathbf V_{jk}\gets\mathbf V_{jk}\cdot\frac{\sqrt{\sqrt{\Delta_{jk}}-\lbrack\mathbf V\mathbf U^{T}\mathbf U+\lambda \mathbf{V}\rbrack_{jk}}}{\sqrt{4\alpha[\mathbf{VV}^{T}\mathbf{V}]_{jk}}},\\
		&\Delta_{jk} \! =\! \lbrack\! \mathbf V\mathbf U^{T}\mathbf U+\lambda \mathbf{V}\rbrack^{2}_{jk}\\
		&+\! [8\alpha\mathbf{VV}^{T}\mathbf{V}]_{jk}[2\alpha\mathbf{YV}\! +\!\mathbf F^{T}\mathbf U+\lambda \mathbf{XC}]_{jk}.
	\end{aligned}
\end{equation}

\subsubsection{Optimizing $\mathbf U$, $\mathbf P^i$, and $\mathbf X$}
Similarly, we may derive the updating rules for $\mathbf U$, $\mathbf P^i$, and $\mathbf X$.
Here, we directly present the corresponding rules due to space limitations.
The updating rule for $\mathbf U$ is:
\begin{equation}\label{rule-u}
	\begin{aligned}
		\mathbf U_{jk}\gets\mathbf U_{jk}\cdot\frac{[\mathbf{FV}+\mathbf{PW}]_{jk}}{[\mathbf U\mathbf V^T\mathbf V+\mathbf U]_{jk}}.
	\end{aligned}
\end{equation}
The updating rule for $\mathbf P^i$ is:
\begin{equation}\label{rule-p}
	\begin{aligned}
		\mathbf P^i_{js}\gets\mathbf P^i_{js}\cdot\frac{[\mathbf {H}^i \mathbf {X}+\mathbf U^i\mathbf W^T]_{js}}{[\mathbf P^i\mathbf X^T\mathbf X+\mathbf P^i\mathbf W\mathbf W^T]_{js}},
	\end{aligned}
\end{equation}
where $\mathbf U^i$ stands for the rows in $\mathbf U$ that are related to the features from view $i$, i.e., $\mathbf F^i$.
The updating rule for $\mathbf X$ is:
\begin{equation}\label{rule-x}
	\begin{aligned}
		\mathbf X_{ns}\gets\mathbf X_{ns}\cdot\frac{\sum_{i}[\mathbf {H}^{iT}\mathbf {P}^i]_{ns}+\lambda[\mathbf V\mathbf C^T]_{ns}}{\lambda[\mathbf X\mathbf {CC}^T]_{ns}+\sum_{i}[\mathbf X\mathbf P^{iT}\mathbf P^i]_{ns}}.
	\end{aligned}
\end{equation}

\subsubsection{Optimizing $\mathbf C$ and $\mathbf W$}
Directly optimizing $\mathbf C$ or $\mathbf W$ is intractable as there are discrete values in each row of $\mathbf C$ or $\mathbf W$.
However, they can be optimized by relaxing the constraints.
To optimize $\mathbf C$ and $\mathbf W$, the corresponding terms in Eq.~(\ref{obj}) can be further relaxed as follows:
\begin{equation}
	\begin{aligned}
		&\text{minimize}\\
		&O(\mathbf C, \mathbf W)=\!\left\Vert \mathbf P\mathbf W\!-\!\mathbf U\right\Vert ^{2}_F+\lambda\left\Vert \mathbf X\mathbf C\!-\!\mathbf V\right\Vert ^{2}_F\\
		&+\delta[\left\Vert \mathbf C\mathbf C^T\!-\!\mathbf I\right\Vert ^{2}_F+\left\Vert \mathbf W\mathbf W^T\!-\!\mathbf I\right\Vert ^{2}_F],\\
		&\text{subject to } \mathbf C\ge0,\mathbf W\ge0,
	\end{aligned}
\end{equation}
where $\delta$ is a penalty parameter. The discreteness in each row of $\mathbf C$ or $\mathbf W$ is guaranteed when $\delta$ is set to be large enough, e.g., $10^5$.
Denoting $\mathbf{\eta}_{sk}$ as the Lagrange multiplier for $\mathbf C_{sk}\ge$ 0, we may construct the Lagrange function for $\mathbf C$:
\begin{equation}\label{langran-c}
	L(\mathbf{C},\mathbf{\eta})=O(\mathbf{C})-\mathit{tr}(\mathbf{\eta}^T\mathbf C).
\end{equation}
According to the KKT conditions, we have the following equation system, based on which we can derive the updating rule for $\mathbf C$:
\begin{equation}\label{kkt-c}
	\begin{aligned}
		&\frac{\partial L}{\partial \mathbf C_{sk}}=2\lambda[-\mathbf {X}^T\mathbf {V} +\mathbf X^T\mathbf X\mathbf C]_{sk}\\
		&+4\delta[ \mathbf C\mathbf C^T\mathbf C-\mathbf C]_{sk}-\eta_{sk}=0,\\
		&\mathbf{\eta}_{sk}\cdot\mathbf{C}_{sk}=0, \mathbf{\eta}_{sk}\ge0, \mathbf{C}_{sk}\ge0.
	\end{aligned}
\end{equation}
The updating rule for $\mathbf C$ can then be obtained by solving Eq.~(\ref{kkt-c}):
\begin{equation}\label{rule-c}
	\begin{aligned}
		&\mathbf C_{sk}\gets\mathbf C_{sk}\cdot\frac{\sqrt{\sqrt{\Delta_{sk}}-\lbrack\lambda\mathbf X^{T}\mathbf X\mathbf C\rbrack_{sk}}}{\sqrt{4\delta[\mathbf{CC}^{T}\mathbf{C}]_{sk}}},\\
		&\Delta_{sk} \! =\! \lbrack\! \lambda\mathbf X^{T}\mathbf X\mathbf C\rbrack^{2}_{sk}\!+\! [8\delta\mathbf{CC}^{T}\mathbf{C}]_{sk}[2\delta\mathbf{C}\! +\lambda\!\mathbf X^{T}\mathbf V]_{sk}.
	\end{aligned}
\end{equation}
Similarly, we can derive the updating rule for $\mathbf W$:
\begin{equation}\label{rule-w}
	\begin{aligned}
		&\mathbf W_{sk}\gets\mathbf W_{sk}\cdot\frac{\sqrt{\sqrt{\Delta_{sk}}-\lbrack\mathbf P^{T}\mathbf P\mathbf W\rbrack_{sk}}}{\sqrt{4\delta[\mathbf{WW}^{T}\mathbf{W}]_{sk}}},\\
		&\Delta_{sk} \! =\! [[\mathbf P^{T}\mathbf P\mathbf W\rbrack^{2}\!+\! [8\delta\mathbf{WW}^{T}\mathbf{W}][2\delta\mathbf{W}\! +\!\mathbf P^{T}\mathbf U]]_{sk}.
	\end{aligned}
\end{equation}
Through iteratively updating $\mathbf V$,  $\mathbf U$, $\mathbf P^i$, $\mathbf X$, $\mathbf C$, and $\mathbf W$ based on Eqs. (\ref{rule-v}),  (\ref{rule-u}), (\ref{rule-p}), (\ref{rule-x}), (\ref{rule-c}), and (\ref{rule-w}) respectively, $O$ in Eq.~(\ref{obj}) can converge in a finite number of iterations.
The details of model optimization are summarized in Algorithm~1.

\begin{algorithm}

	\caption{GCCFP}
	\KwIn{Graph Data: $\mathbf Y$, $\mathbf {F} = [\mathbf {F}^1;...\mathbf {F}^D]$}
	\KwOut{$\mathbf V$, $\mathbf U$, $\mathbf P^i$, $\mathbf X$, $\mathbf C$, $\mathbf W$;}
	Compute $\mathbf D$ by Eq. (\ref{sdiffusion}), $\mathbf Y = \mathbf D$;\\
	Compute $\lbrace\mathbf H^{i}\rbrace^{D}_{i=1}$ by Eq.~(\ref{feature-propagation});\\
	Initialize $\mathbf V$, $\mathbf U$, $\mathbf X$, $\mathbf C$, $\mathbf W$, $\lbrace\mathbf P^{i}\rbrace^{D}_{i=1}$;\\
	$l \leftarrow 0$;\\
	\While{$l < T_{max}$}
	{
		$l \leftarrow l+1$;\\
		Update $\mathbf V$ and $\mathbf U$ by Eqs.~(\ref{rule-v}) and (\ref{rule-u});\\
		
		Update $\lbrace\mathbf P^{i}\rbrace^{D}_{i=1}$ by Eq.~(\ref{rule-p});\\
		Update $\mathbf X$, $\mathbf C$ and $\mathbf W$ by Eqs.~(\ref{rule-x}),~(\ref{rule-c}), and (\ref{rule-w});\\
		
		Calculate objective value $O^{(l)}$ by Eq.~(\ref{obj});\\
		\If{$O^{(l-1)}-O^{(l)}\le \epsilon$}{break;\\}
	}
	Extracting cluster membership based on $\mathbf V$;\\
	
	\label{DMVFPPGC-algorithm}
\end{algorithm}

\section{Analysis on model complexity}\label{complexity}
Given the updating rules presented in Eqs.~(\ref{rule-v}), (\ref{rule-u}), (\ref{rule-p}), (\ref{rule-x}), (\ref{rule-c}), and (\ref{rule-w}), the computational complexity of GCCFP can be analyzed as follows.
Given Eq.~(\ref{rule-v}), we can derive that the complexity of updating of all latent variables in $\mathbf V$ is about $O(2NMK+2NK^2+N^2K^2+N^2K+NSK)$.
Similarly, we can derive the computational complexity of updating other latent variables.
Updating $\mathbf U$, $\mathbf P^i$, $\mathbf X$, $\mathbf C$, and $\mathbf W$ follows the order of $O(MNK+MSK+MNK^2+MK^2)$, $O(M^iNS+M^iSK+M^iNS^2+M^iS^2K+2M^iS^2)$, $O(2NMS+3NS^2+NS^2K)$, $O(S^2K^2+SK^2+NS^2K+S^2K+NSK)$, and $O(MS^2K+S^2K+S^2K^2+SK^2+MSK)$, respectively.
Overall, the computational complexity of GCCFP is approximately $O(N^2K^2+NMK^2+NMS^2)$.

\begin{table}[b]
	\centering
	\normalsize\caption{Statistics of the test datasets. A social, document, or biological graph is abbreviated as Soc, Doc, or Bio.
		}
	\label{data}    
	\small\centering\begin{tabular}{c p{0.45cm} c c c c p{0.45cm}}
		\hline\hline
		Dataset &Type& $N$&$|E|$&$M$&$D$&$K$\\
		\noalign{\smallskip}\hline\noalign{\smallskip}
		Ego&Soc&4039&88234&1283&1&191\\
		Twitter&Soc&3687&49881&20905&2&242\\
		Gplus&Soc&107614&3755989&13966&5&463\\\hline
		Cornell&Doc&195&283&1588&1&5\\
		Texas&Doc&187&280&1501&1&5\\
		Washington&Doc&230&366&1579&1&5\\
		Wisconsin&Doc&265&459&1626&1&5\\
		Wiki&Doc&2405&17981&4973&1&17\\\hline
		DIP&Bio&4579&20845&4237&3&200\\
		Biogrid& Bio  & 5640   & 59748  & 4286 &3 & 200  \\
		\hline\hline
	\end{tabular}
\end{table}

\begin{table*}
	\centering
	\caption{Performance evaluation using $NMI$. The best performance on each dataset is highlighted in bold.}
	\resizebox{\textwidth}{!}{%
			\begin{tabular}{c|c|ccc|ccccc|cc}
				\hline\hline
				\multicolumn{2}{c|}{\diagbox{\small{Approaches}}{\small{Datasets}}}                                                      & Ego & Twitter & Gplus & Cornell & Texas & Washington & Wisconsin & Wiki & Dip & Biogrid \\ \hline
				\multicolumn{1}{c|}{\multirow{14}{*}{Shallow}} & 
                \multirow{2}{*}{CoDA}     &55.505&60.878&18.900&24.904&16.730&22.174&8.913&30.203&76.314&73.229         \\
				\multicolumn{1}{c|}{}                          &                           &$\pm$1.066&$\pm$0.559&$\pm$1.044&$\pm$0.131&$\pm$1.727&$\pm$0.778&$\pm$0.213&$\pm$0.236&$\pm$0.707&$\pm$0.601         \\ \cline{2-12} 
				\multicolumn{1}{c|}{}                          & \multirow{2}{*}{SBM}      &63.527&12.276&9.871&7.472&4.361&12.838&13.303&22.670&82.497&88.868         \\
				\multicolumn{1}{c|}{}                          &                           &$\pm$0.403&$\pm$1.505&$\pm$0.437&$\pm$1.731&$\pm$0.239&$\pm$0.007&$\pm$0.239&$\pm$0.008&$\pm$0.825&$\pm$0.734         \\ \cline{2-12} 
				\multicolumn{1}{c|}{}                          & \multirow{2}{*}{HPNMF}    &59.954     &63.272         &17.086       &9.576         &  8.506     & 15.402           &  17.239         & 27.550     & 86.641    &  90.636       \\
				\multicolumn{1}{c|}{}                          &                          &$\pm$0.435     &  $\pm$ 0.619       &$\pm$0.698       & $\pm$0.613        &  $\pm$0.381      &  $\pm$0.595          &   $\pm$0.935        &  $\pm$0.913    &$\pm$0.996     &  $\pm$0.844       \\ \cline{2-12} 
				\multicolumn{1}{c|}{}                          & \multirow{2}{*}{CESNA}    &57.513&46.588&10.164&17.069&9.255&22.177&16.482&9.374&77.286&80.088         \\
				\multicolumn{1}{c|}{}                          &                           &$\pm$1.119&$\pm$0.155&$\pm$0.601&$\pm$0.157&$\pm$0.106&$\pm$0.536&$\pm$0.565&$\pm$0.185&$\pm$0.148&$\pm$0.353         \\ \cline{2-12} 
				\multicolumn{1}{c|}{}                          & \multirow{2}{*}{SCI}      &44.025&56.889&10.241&9.809&14.917&6.912&12.848&22.012&90.927&92.028         \\
				\multicolumn{1}{c|}{}                          &                           &$\pm$0.509&$\pm$0.196&$\pm$0.775&$\pm$0.962&$\pm$6.419&$\pm$0.680&$\pm$1.046&$\pm$0.536&$\pm$0.230&$\pm$0.195         \\ \cline{2-12} 
				\multicolumn{1}{c|}{}                          & \multirow{2}{*}{MISAGA}   &56.452&65.329&10.245&26.773&27.649&\bf39.177&31.004&38.989&87.952&89.485         \\
				\multicolumn{1}{c|}{}                          &                           &$\pm$0.538&$\pm$0.226&$\pm$1.158&$\pm$5.639&$\pm$0.321&$\pm$2.925&$\pm$5.842&$\pm$0.523&$\pm$0.223&$\pm$0.314         \\ \hline
				\multicolumn{1}{c|}{\multirow{4}{*}{Deep}}     & \multirow{2}{*}{VGAE}     &  60.088   &  59.873       &15.849       &8.116         &  8.693     &   9.416         &6.511           & 18.590     &83.315     & 85.604        \\
				\multicolumn{1}{c|}{}                          &                           &  $\pm$0.895   &  $\pm$0.961       &  $\pm$0.940     & $\pm$0.493        & $\pm$0.284      & $\pm$0.479           &$\pm$0.242           &$\pm$0.339      &$\pm$0.639     &  0.386       \\ \cline{2-12} 
				\multicolumn{1}{c|}{}                          & \multirow{2}{*}{ARVGAE}   &60.904     &  64.952       &  18.439     & 8.979        &  10.108     &  10.525          &  13.269         & 34.724     &88.755     & 89.720        \\
				\multicolumn{1}{c|}{}                          &                           & $\pm$0.709    &  $\pm$0.887       &  $\pm$0.816     & $\pm$0.654        & $\pm$0.485      & $\pm$0.578           & $\pm$0.807          &  $\pm$0.941    &$\pm$0.796     & $\pm$0.611        \\ \hline
				& \multirow{2}{*}{GCCFP} &\bf67.193&\bf68.653&\bf23.046&\bf29.715&\bf33.995&35.865&\bf46.962&\bf46.317&\bf91.544&\bf93.135         \\
				&                           &$\pm$0.649&$\pm$1.121&$\pm$0.846&$\pm$0.823&$\pm$0.933&$\pm$0.597&$\pm$0.938&$\pm$0.467&$\pm$0.104&$\pm$0.219         \\ \hline\hline
			\end{tabular}%
		}
		\label{nmi}
	\end{table*}

\section{Experiments and analysis}\label{exp}
In this section, we comprehensively evaluate the effectiveness of GCCFP by comparing it with prevalent approaches to graph clustering on diverse real-world tasks.

\subsection{Experimental set-up}
\subsubsection{Datasets}
Ten real-world datasets with ground-truth clusters are used in our experiment to verify the effectiveness of different approaches. They include three social graphs, which are $Ego-facebook$ \cite{yang2013}, $Twitter$ \cite{yang2013}, and $Google+$ \cite{yang2013}; five document citation graphs, which are $Cornell$ \cite{lu2003}, $Texas$ \cite{lu2003}, $Washington$ \cite{lu2003}, $Wisconsin$ \cite{lu2003}, and $Wiki$ \cite{lu2003}; and two biological graphs, $DIP$ \cite{xenarios2002dip} and $Biogrid$ \cite{stark2006biogrid}.
As the size and the number of views w.r.t. vertex features are different from one to another, these datasets may well reveal the robustness of different approaches.
We have summarized the characteristics of all the test datasets in Table~\ref{data}.

\subsubsection{Baselines and evaluation metrics}
Eight approaches to graph clustering, including CoDA \cite{yang2014detecting}, SBM \cite{peng2015scalable}, HPNMF \cite{ye2019homophily}, CESNA \cite{yang2013}, SCI \cite{wang2016semantic}, MISAGA \cite{he2017misaga}, VGAE \cite{kipf2016variational}, and ARVGAE \cite{pan2018adversarially} are selected as baselines for comparison.
These baselines are representative approaches based on graph structure, designed for AG clustering, or developed with graph neural networks.
For fair comparisons, we use the official source codes of all baselines that are publicly available and tune these baselines according to the recommended settings.
To evaluate the clustering performance of all approaches, we consider using Normalized Mutual Information ($NMI$) as the metric.
All experiments were conducted on a workstation with a 6-core 3.4GHz CPU and 32GB RAM. 
Each method was executed ten times on each dataset, and the average performance was reported.

\subsection{Clustering performance in real-world graphs}
The average clustering performance evaluated by $NMI$ has been listed in Table \ref{nmi}.
Notably, the GCCFP model showcases its adaptability by securing the top performance on nine testing datasets, and even on the challenging $Washington$ dataset, it ranks second.
The proposed model can obtain at least 5\% performance improvement compared with other baselines on seven datasets.
Specifically, GCCFP outperforms SBM by 5.77\% on $Ego-facebook$.
On the $Google+$ dataset, the proposed approach is better than CoDA by 21.94\%.
On $Twitter$, $Cornell$, $Texas$, $Wisconsin$, and $Wiki$, GCCFP may outperform MISAGA by 5.09\%, 10.99\%, 22.95\%, 51.47\%, and 18.79\%, respectively.
Given the experimental results in terms of $NMI$, GCCFP is a robust method for uncovering clusters in different types of AGs constructed based on real-world data.
The clustering results presented in Table \ref{nmi} demonstrate that GCCFP is a very effective method for uncovering clusters in diverse real-world graphs.

\begin{figure}[htbp]
	\centering
	\includegraphics[width=\columnwidth]{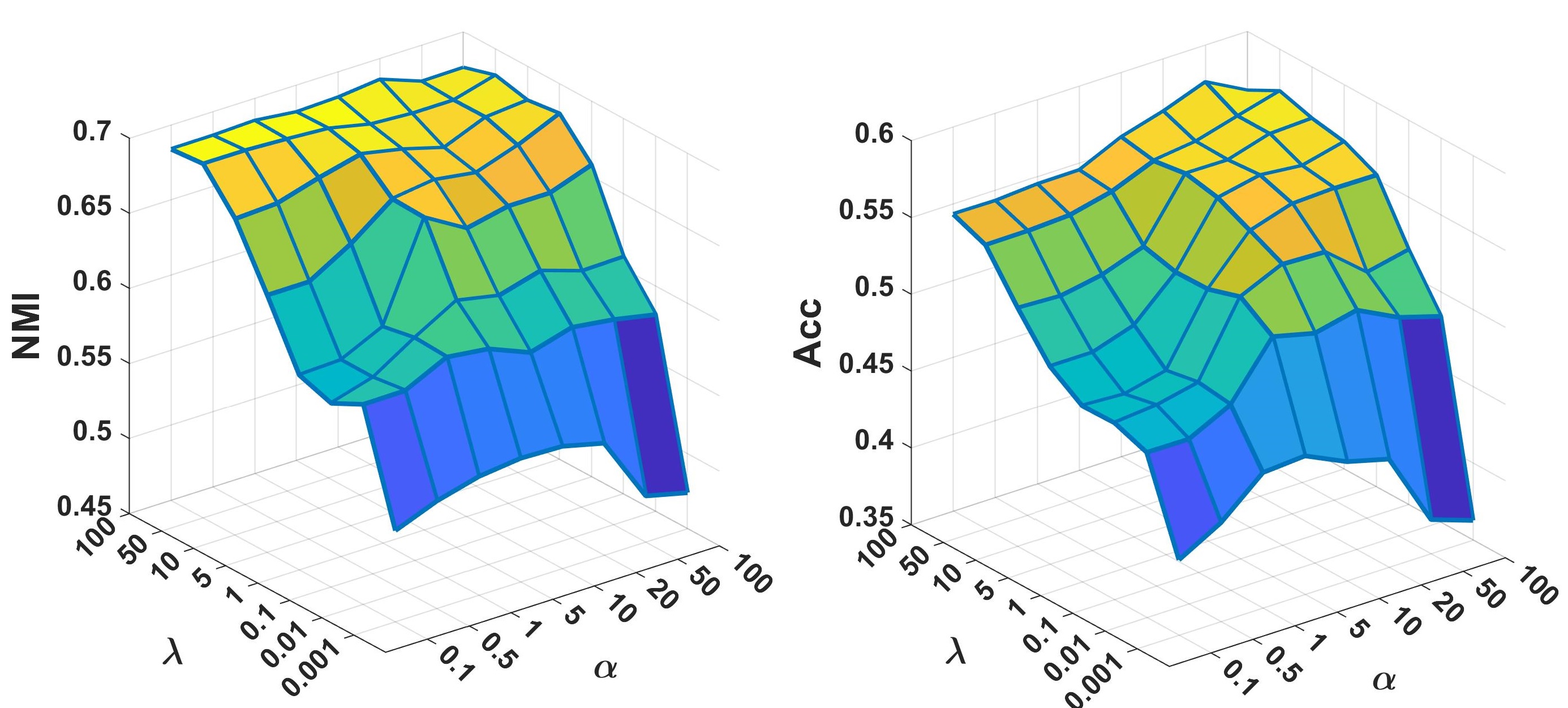}
	\vspace{-0.55cm}\caption{
		Sensitivity test of hyper-parameters $\alpha$ and $\lambda$ on $Ego-facebook$. 
		}
	\label{sens}
\end{figure}

\subsection{Sensitivity test of hyper-parameters}
As previously introduced, $\alpha$ and $\lambda$ are adopted by the proposed GCCFP to balance the relative significance of global feature and structural propagation modeling and regularize multi-view latent propagation to influence the learning of cluster membership.
In this subsection, we investigate how these two hyper-parameters can affect the clustering performance of GCCFP.
Specifically, we let GCCFP discover clusters on all test datasets by setting $\alpha = [0.1, 0.5, 1, 5, 10, 20, 50, 100]$ and $\lambda = [0.001, 0.01, 0.1, 1, 5, 10, 50, 100]$.
We then evaluate the clustering performance of GCCFP using $NMI$ and $Acc$.
For example, the results collected from $Ego-facebook$ are depicted in Fig.~\ref{sens}.
As observed, GCCFP can robustly under a broad combination of $\alpha$ and $\lambda$.
Moreover, GCCFP can obtain better $NMI$ and $Acc$ when $\alpha$ and $\lambda$ are set as larger values, e.g., $\alpha \geq 5$ and $\lambda \geq 1$.
Thus, in our experiments, we let GCCFP perform clustering tasks by setting $\alpha = 5$ and $\lambda =1$.

\subsection{Results of model convergence}\label{convegence-time}
\begin{figure}[t]
	\centering
	\includegraphics[width=\columnwidth]{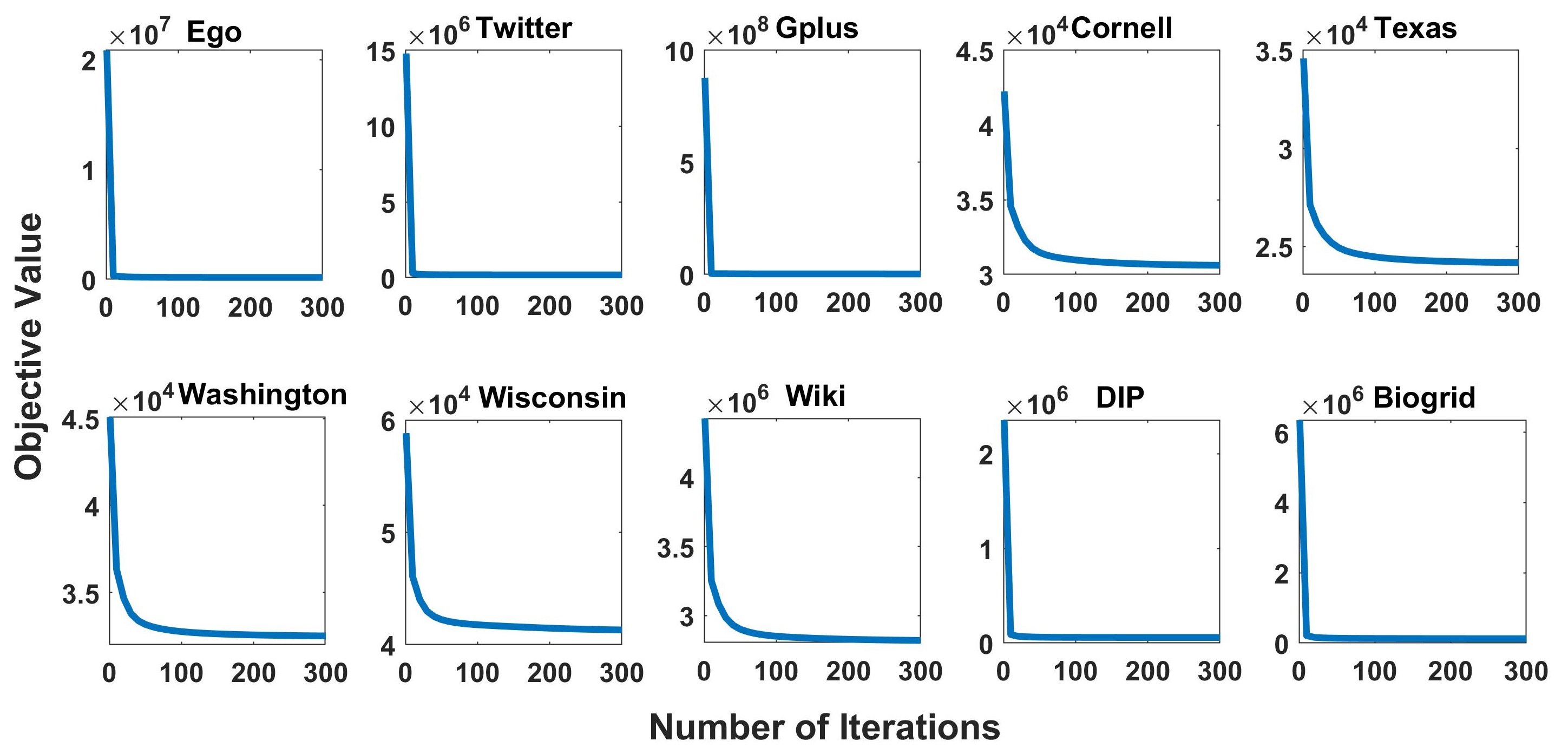}
	\caption{Model convergence on testing datasets.}\label{convergence}
\end{figure}

In this subsection, we investigate the convergent speed of GCCFP on all testing datasets.
Specifically, we ran GCCFP for 300 iterations in each testing dataset and recorded the objective values, which have been depicted as curves in Fig. \ref{convergence}.
As shown by the figure, the value of the objective function can converge to a stable state in fewer than 300 iterations on all ten testing datasets.
These results indicate that GCCFP is capable of learning accurate clustering results efficiently.

\section{Conclusion}\label{conclusion}
In this paper, we have proposed a novel model, namely Graph Clustering With Cross-View Feature Propagation (GCCFP).
Unlike existing approaches, GCCFP further considers leveraging the pivotal latent propagation of multi-view vertex features to regularize the learning process of cluster membership for each vertex in the graph.
The cluster membership learned by GCCFP is jointly determined by graph topology, global vertex features, and corss-view feature propagation. Therefore, GCCFP can uncover meaningful clusters in various types of graph data.
Having been extensively tested on several real-world graphs and compared with both classical and prevalent methods for graph clustering, GCCFP is demonstrated to be capable of accurately and efficiently discovering graph clusters.
In the future, we will improve the efficiency of GCCFP by designing distributed or stochastic algorithms for model optimization.
In addition, we will improve the effectiveness of the proposed model by enabling it to be concerned with the cross-view propagation effect of cross-view vertex features.

\balance
\bibliographystyle{IEEEtran}
\bibliography{ref}

\end{document}